\documentclass{article}
\usepackage{arxiv}

\usepackage[utf8]{inputenc} 
\usepackage[T1]{fontenc}    
\usepackage{hyperref}       
\usepackage{url}            
\usepackage{booktabs}       
\usepackage{amsfonts}       
\usepackage{nicefrac}       
\usepackage{microtype}      
\usepackage{lipsum}		
\usepackage{graphicx}
\usepackage{natbib}
\usepackage{doi}

\usepackage{graphicx}
\usepackage{multirow}
\usepackage{todonotes}
\usepackage{comment}
\usepackage{adjustbox}
\usepackage{amsmath}
\usepackage[utf8]{inputenc}
\usepackage[english]{babel}
\usepackage{color}
\usepackage{amsfonts}
\usepackage{url}
\usepackage[mathscr]{euscript}
 \let\mathscr\relax%
\usepackage[scr]{rsfso}
\usepackage{caption}
\usepackage[labelfont=bf]{caption}

\usepackage{tabularx,booktabs}
\newcolumntype{C}{>{\centering\arraybackslash}X} 
\usepackage{multicol}
\usepackage{algorithm2e}
\usepackage{booktabs,ragged2e}
\usepackage[utf8]{inputenc}
\usepackage{todonotes}
\usepackage{lineno}
\usepackage{makecell}
\usepackage[flushleft]{threeparttable}
\usepackage{subcaption}
\usepackage{color}

\usepackage[nolist,nohyperlinks]{acronym}
\usepackage{footnote}
\usepackage[perpage]{footmisc}
\usepackage[autostyle]{csquotes}
\makesavenoteenv{tabular}
\acrodef{DL}{Deep learning}
\acrodef{CNN}{Convolutional Neural Network}
\acrodef{ML}{Machine Learning}
\acrodef{mIoU}{mean Intersection over Union}
\acrodef{GI}{gastrointestinal}
\acrodef{AI}{Artificial Intelligence} 
\acrodef{CADx}{computer aided diagnosis} 
\acrodef{CRC}{colorectal cancer}
\acrodef{DSC}{Dice Coefficient}
\acrodef{mDSC}{Dice Coefficient}
\acrodef{OOD}{Out-Of-Distribution}
\acrodef{SOTA}{State-of-the-art}
\acrodef{HD}{Hausdorff distance}

\newcommand{\equalcontribA}{\textsuperscript{\dag}}
\newcommand{\equalcontribB}{\textsuperscript{\ddag}}

\title{A Novel Momentum-Based Deep Learning Techniques for Medical Image Classification and Segmentation}

\date{} 					

\author{ \href{https://orcid.org/0000-0002-9818-8966}{\includegraphics[scale=0.06]{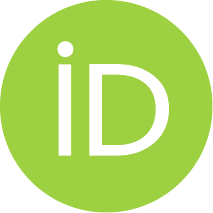}\hspace{1mm}Koushik Biswas}\equalcontribA \\
	Machine and Hybrid Intelligence Lab\\
	Northwestern University\\
	USA \\
	\texttt{koushik.biswas@northwestern.edu} \\
	\And
	\href{https://orcid.org/0000-0003-1561-1173}{\includegraphics[scale=0.06]{orcid.pdf}\hspace{1mm}Ridam Pal}\equalcontribA \\
	IIIT Delhi\\
	New Delhi, India \\
	\texttt{ridamp@iiitd.ac.in} \\
        \And
	\href{https://orcid.org/0000-0003-2798-8672}{\includegraphics[scale=0.06]{orcid.pdf}\hspace{1mm}Shaswat Patel}\equalcontribB \\
	NSIT Delhi\\
	New Delhi, India \\
	\texttt{shaswat178@gmail.com} \\
       \And
       \href{https://orcid.org/0000-0002-8078-6730}{\includegraphics[scale=0.06]{orcid.pdf}\hspace{1mm}Debesh Jha}\equalcontribB \\
	Machine and Hybrid Intelligence Lab\\
	Northwestern University\\
	USA \\
	\texttt{debesh.jha@northwestern.edu} \\
        \And
         \href{https://orcid.org/0000-0002-0440-3785}{\includegraphics[scale=0.06]{orcid.pdf}\hspace{1mm}Meghana Karri}\equalcontribB \\
	Machine and Hybrid Intelligence Lab\\
	Northwestern University\\
	USA \\
	\texttt{meghana.karri@northwestern.edu} \\
        \And
       \href{https://orcid.org/0000-0001-7934-0259}{\includegraphics[scale=0.06]{orcid.pdf}\hspace{1mm}Amit Reza}\\
	Space Research Institute (IWF) of Austrian Academy of Sciences\\
	Graz, Austria\\
	\texttt{amitreza@gmail.com} \\
        \And
        {Gorkem Durak} \\
	Machine and Hybrid Intelligence Lab\\
	Northwestern University\\
	USA \\
	\texttt{gorkem.durak@northwestern.edu} \\
         \And
        {Alpay Medetalibeyoglu}\\
	Machine and Hybrid Intelligence Lab\\
	Northwestern University\\
	USA \\
	\texttt{medetalibeyoglu.alpay@northwestern.edu} \\
        \And
        {Matthew Antalek}\\
	Machine and Hybrid Intelligence Lab\\
	Northwestern University\\
	USA \\
	\texttt{matthew.antalek@nm.org} \\
        \And
        {Yury Velichko}\\
	Machine and Hybrid Intelligence Lab\\
	Northwestern University\\
	USA \\
	\texttt{y-velichko@northwestern.edu} \\ 
        \And
        \href{https://orcid.org/0000-0001-5526-8272}{\includegraphics[scale=0.06]{orcid.pdf}\hspace{1mm}Daniela Ladner} \\
	Northwestern University\\
	USA \\
	\texttt{dladner@northwestern.edu} \\
        \And
        {Amir Borhani}\\
	Northwestern University\\
	USA \\
	\texttt{amir.borhani@northwestern.edu} \\
        \And
        \href{https://orcid.org/0000-0001-7379-6829}{\includegraphics[scale=0.06]{orcid.pdf}\hspace{1mm}Ulas Bagci}\thanks{} \\
	Machine and Hybrid Intelligence Lab\\
	Northwestern University\\
	USA \\
	\texttt{ulas.bagci@northwestern.edu} \\
}

\date{}

\renewcommand{\shorttitle}{Momentum Method for Medical Imaging}

\hypersetup{
pdftitle={Momentum Method for Medical Imaging},
pdfsubject={Computer Vision, Machine Learning},
pdfauthor={K.Biswas et al.},
pdfkeywords={Liver segmentation, Lung Segmentation, Polyp segmentation, Medical Image Classification}}

\begin{document}
\maketitle
\footnotetext[1]{Equal contribution: Koushik Biswas and Ridam Pal.}
\footnotetext[2]{Equal contribution: Shaswat Patel, Debesh Jha, and Meghana Karri.}

\begin{abstract}
	Accurately segmenting different organs from medical images is a critical prerequisite for computer-assisted diagnosis and intervention planning. This study proposes a deep learning-based approach for segmenting various organs from CT and MRI scans and classifying diseases. Our study introduces a novel technique integrating momentum within residual blocks for enhanced training dynamics in medical image analysis. We applied our method in two distinct tasks: segmenting liver, lung, \& colon data and classifying abdominal pelvic CT and MRI scans. The proposed approach has shown promising results, outperforming state-of-the-art methods on publicly available benchmarking datasets. For instance, in the lung segmentation dataset, our approach yielded significant enhancements over the TransNetR model, including a 5.72\% increase in dice score, a 5.04\% improvement in mean Intersection over Union (mIoU), an 8.02\% improvement in recall, and a 4.42\% improvement in precision. Hence,  incorporating momentum led to state-of-the-art performance in both segmentation and classification tasks, representing a significant advancement in the field of medical imaging. 

\end{abstract}





\section{Introduction}

In modern medicine, medical imaging plays an important role in bridging visual data and clinical insights. Computer vision plays a key role in improving the interpretation of complex medical images, including CT, X-rays, and MRIs. This transformational field plays a pivotal role in automating abnormality detection, classifying anatomical structures, and quantifying disease features. Real-time surgical guidance, minimally invasive procedures, and ongoing research into interpretability and deployment further emphasize its significance. Healthcare professionals gain the ability to make informed decisions based on a deeper understanding of patients’ conditions, even at the early stages. This transformative capability encourages collaboration among computer scientists, clinicians, and researchers, driving innovations with great promise for healthcare outcomes.

Deep learning has emerged as a significant tool in clinical support for medical imaging, revolutionizing disease detection, segmentation, and classification~\cite{review}. By automating feature extraction, deep learning models, particularly convolutional neural networks (CNNs), have demonstrated the ability to learn hierarchical features from raw medical images. Techniques such as segmentation and classification enable these models to extract critical visual cues from various imaging modalities, including CT, MRI, and endoscopy. These advancements assist clinicians in disease staging, surgical planning, and assessing treatment responses. Furthermore, deep learning is good at analyzing patterns in extensive datasets, which can aid in early disease detection. 

However, medical images are more complex than standard images, which makes analyzing them thoroughly quite challenging. Researchers have been working hard to solve these challenges, especially in areas like early diagnosis and quantitative imaging, where mistakes can be really risky. Colorectal cancer is one of the most common cancers worldwide. Detecting polyps early is crucial, as some types can develop into cancer if not addressed at an early stage. However, sometimes, it's tough for doctors to differentiate polyps from normal tissue visually. Fortunately, deep learning (DL) models have become a powerful tool for identifying and categorizing abnormalities from medical images. 


The introduction of residual connection and self-attention mechanisms has further enhanced deep learning and computer vision domain, allowing them to focus on crucial clinical regions within an image and leading to more robust architectures~\cite{review1,review2}. These innovative methods have greatly enhanced medical imaging, leading to more efficient diagnoses and streamlined workflows, thus reducing the strain on healthcare professionals and clinical resources over the past decade.

Our study introduces a method that utilizes the power of the \textbf{momentum} term within the design of the residual block. Incorporating momentum within the residual blocks offers effective network training to enhance the learning algorithm. Our experiments demonstrated that this enhancement led to faster convergence and improved stability, which had the potential to achieve superior performance. The efficacy of our proposed method is supported by extensive evaluations across various tasks, including lung, liver, and polyp segmentation, as well as the classification of abdominal pelvic CT and MRI scans (on  RadImageNet Data). Based on these extensive experiments, it is clear that including a momentum-based method in residual block outperforms the current state-of-the-art methods.

\section{Related Works and Motivation}
Accurate segmentation and classification of medical images are crucial for com-puter-aided diagnosis and treatment planning but remain challenging due to factors like low contrast, noise, and patient variability. Convolutional neural networks (CNNs) have shown promise in these tasks, automating diagnosis and aiding medical decision-making~\cite{alexnet,vgg}. With advancements in deep learning, skip-connection was crucial in addressing the degradation problem arising from vanishing gradients. Introducing skip connections in architectures like ResNet has addressed the vanishing gradient problem, improving the training of deep networks~\cite{resnet}. The use of residual connections has become widespread across a variety of image classification models, including CapsuleNet~\cite{capsule}, PreactResNet~\cite{preact}, MobileNet V2~\cite{mobilenetv2}, and ShuffleNet~\cite{shufflenet}, among others. Drozdzal et al.~\cite{importance} demonstrated that incorporating long and short skip connections in Fully Convolutional Networks (FCNs) enhances biomedical image segmentation without additional post-processing. U-Net's architecture captured high-level features and helped in reconstructing the segmentation map \cite{unet}. The drawbacks have been further improved by ResUNet, combining the strengths of Residual Networks and U-Net for better performance~\cite{resunet}. These advancements highlight the critical role of deep learning in enhancing medical image analysis. In addition, this connection has been extensively utilized in many segmentation models, such as TransNetR~\cite{transnetr}, ResUNet++~\cite{resunetplus}, and PVTformer~\cite{pvt}, among others. Due to its versatility and effectiveness, it has become a crucial component in modern deep-learning models, especially in computer vision and medical imaging applications.

Building on previous research, we have integrated the momentum term into ResNet blocks for both segmentation and classification tasks across various models. For segmentation, we have integrated into architectures like UResNet and ResUnet++, while for image classification, we used models such as MobileNet and ShuffleNet. Our study aims to enhance training dynamics, improve generalization, boost performance metrics, and increase the efficiency and scalability of deep neural networks. The addition of momentum can improve convergence, and potentially lead to state-of-the-art results in segmentation and classification tasks, offering a novel contribution to the field of medical imaging with an impact on future neural network architectures and training strategies. Our contributions to this paper are as follows: 

\begin{enumerate}
    \item We have integrated the momentum term in the resnet block in various neural network architectures for both segmentation and classification tasks.
    \item Our extensive experiments on large datasets demonstrate that our proposed momentum-based architecture significantly enhances the ability of previous models to identify complex data patterns, leading to more accurate predictions on unseen data or test datasets.
    
\end{enumerate}

\section{Method}{\label{section:method}}
Deep convolutional neural networks have demonstrated exceptional performance in various computer vision tasks and have become state-of-the-art in image classification problems. AlexNet \cite{alexnet}, VGG \cite{vgg}, ResNet \cite{resnet}, Vision Transformer \cite{vit} are some popular architectures for image classification problem. However, as the number of layers increases, the problem of vanishing gradients becomes increasingly prevalent, which leads to a drop in training accuracy beyond a certain depth. We present our approach in the next subsection. 

\subsection{Residual Network}
In 2015, a group of researchers from Microsoft Research introduced ResNet \cite{resnet}. This architecture is designed to overcome the problem of degradation that deep neural networks often face. As the number of layers increases, the accuracy of the network can either saturate or degrade. ResNet addresses this issue by introducing residual connections, also known as skip connections. These connections allow the network to bypass one or more layers during training and inference. Residual connections work by adding the output of a layer to the output of a few layers ahead, creating a shortcut path for gradient flow. This enables the training of much deeper networks. ResNets have been widely adopted in various computer vision tasks such as image classification, object detection, and semantic segmentation. They have achieved state-of-the-art performance on several benchmark datasets.

\subsection{Proposed Momentum-based Approach for medical images}
\begin{figure}
   \begin{center}
  \makebox[\textwidth]{\includegraphics[width=7cm]{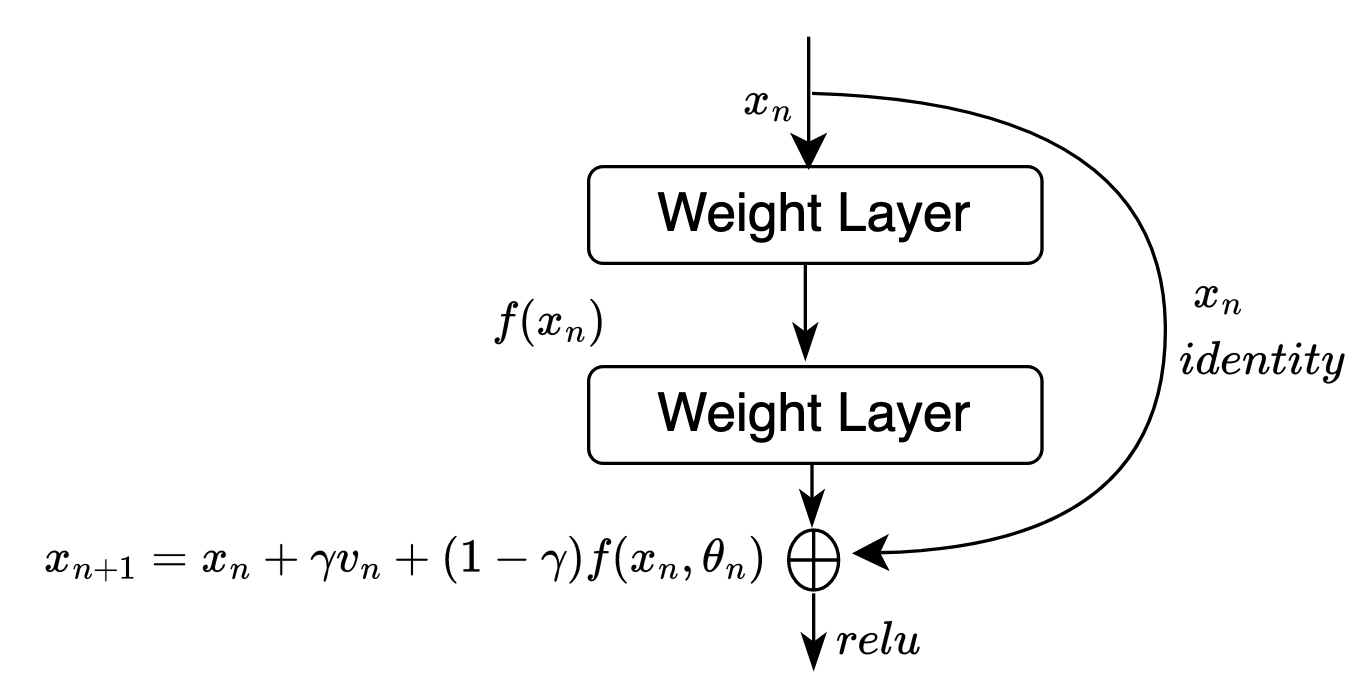}}
\end{center}
    \caption{An integration of the momentum term in the ResNet Block}
    \label{fig:enter-label}
\end{figure}

The momentum ResNet~\cite{momentum} relies on the integration of the momentum within the Residual Block for image classification, as shown in Figure~\ref{fig:enter-label}. Our methodology is designed to optimize network performance in medical image classification and semantic segmentation tasks. The feed-forward residual block at layer $n$ is defined as follows:

\begin{align}
    x_{n+1} = x_n + f(x_n, \theta_{n})
\end{align}
The velocity equation is defined as follows:
\begin{align}
    v_{n+1} = \gamma v_n + (1 - \gamma)f(x_n, \theta_n)
\end{align}
The momentum equation with the residual block is defined as 
\begin{align}
    x_{n+1} = x_n + v_{n+1}
\end{align}
In particular, if we consider $\gamma=0$, we got the classical ResNet architecture, and for $\gamma=1$, we got the RevNet~\cite{reversible} architecture.
\subsection{Reversible Property}
In the realm of deep learning, a neural network is considered reversible if all of its activations can be recalculated when performing a backward pass. By contrast, a network that is not reversible requires saving activations from the forward pass, leading to increased memory usage. Reversible or invertible networks have the unique advantage of being able to perform backpropagation without storing the outputs of activation function, thus significantly reducing the memory footprint of models that employ this approach~\cite{reversible1,reversible,capsulemomentum,momentum}. Momentum-residual block is invertible. We can invert this equation as follows:
\begin{align}
    x_{n} = x_{n+1} - v_{n+1}
\end{align}

\begin{align}
    v_n = \frac{1}{\gamma}(v_{n+1} - (1 - \gamma)f(x_n, \theta_{n})).
\end{align}
\section{Experiments and Results}
We have reported results on medical image segmentation problems on different organs like lungs (medical decathlon data), liver (medical decathlon data), and polyps (Kvasir-SEG dataset). For classification, we have considered RadImageNet data. 
\subsection{Medical Image Segmentation} 
We evaluated the efficacy of our architecture by considering two tasks: segmentation and classification. We consider the decathlon Segmentation Benchmark~\cite{lung,decathelon,liver} and the Kvasir-Seg~\cite{kvasir} datasets for the segmentation task. The Decathlon is a comprehensive collection of medical image segmentation datasets covering various anatomies, modalities, and sources, including the brain, heart, liver, hippocampus, prostate, lung, pancreas, hepatic vessel, spleen, and colon. For our experiments, we consider the Liver~\cite{decathelon,liver} and the Lung~\cite{decathelon,lung} data. The Kvasir-SEG dataset consists of 1000 images, of which 880 were used for training and the remaining for testing. 
\begin{table}[!htbp]
\scriptsize
\centering
\caption{Comparison of different segmentation models and our proposed momentum-based approach on Lung segmentation benchmark dataset.}
\label{tab:segmentationlITS1}
\begin{tabular}{c|c|c|c|c|c|c}
\toprule
\textbf{Model} & \textbf{mDSC} & \textbf{mIoU} & \textbf{Rec.} & \textbf{Prec.} & \textbf{F2}  & \textbf{HD}  \\ \hline

ResUNet++~\cite{resunetplus}  & 32.78 & 25.88 & 35.11 & \textbf{84.29} & 33.88 & 2.66 \\
Momentum-ResUNet++ \textbf{(Ours)}  & \textbf{43.39}  & \textbf{32.32} & \textbf{52.75} & 82.09  & \textbf{46.57} & \textbf{2.25} \\
\hline

ResUNet~\cite{resunet}  & 42.02 & 31.62 & 43.07 & 74.57 & 41.97 &  2.36\\
Momentum-ResUNet \textbf{(Ours)}  & \textbf{43.55} & \textbf{32.35}  & \textbf{44.78} & \textbf{75.45}  & \textbf{42.67} & \textbf{2.34} \\
\hline

TransNetR~\cite{transnetr} & 45.82  & 35.07 & 44.30 & 75.41 & 44.45 & 2.36\\
Momentum-TransNetR \textbf{(Ours)}    & \textbf{51.54}  & \textbf{40.11} & \textbf{52.32} & \textbf{79.83} & \textbf{50.88} & \textbf{2.23} \\
\hline

PVTFormer~\cite{pvt}  & 26.92  & 18.50 & 27.90 & 49.43 & 26.47 & 3.54 \\
Momentum-PVTFormer \textbf{(Ours)}  & \textbf{29.47}  & \textbf{20.70} & \textbf{30.05} & \textbf{55.85} & \textbf{28.88} & \textbf{3.52} \\

\bottomrule
\end{tabular}
\end{table}

To avoid bias, we divided the Liver data into independent training (70 patients), validation (30 patients), and test (30 patients) sets. The volumetric CT scans were processed slice-by-slice to fit into regular computer hardware (GPU). Prior to segmentation, we extracted healthy liver masks for unbiased results.

The liver, Lung, and Kvasir data segmentation experiments were conducted using the PyTorch framework~\cite{pytorch}. We consider a batch size of 16 and a learning rate of 1e$^{-4}$ for the segmentation tasks. We trained the network for 500 epochs with an early stopping patience of 50 to fine-tune the network parameters. To enhance the network performance further, we used a hybrid loss function combining binary cross-entropy and dice loss and an Adam optimizer for updating the parameters. The data was divided into three sets: 80\% for training, 10\% for validation, and 10\% for testing. We resized the image to $256\times256$ pixels in-plane resolution to balance the training time and model complexity. All the segmentation experiments were conducted on the A100 GPU server.
\begin{figure}[!t]
    \centering
    \includegraphics[trim=0cm 18.5cm 0cm 0cm, clip, width=0.6\linewidth]{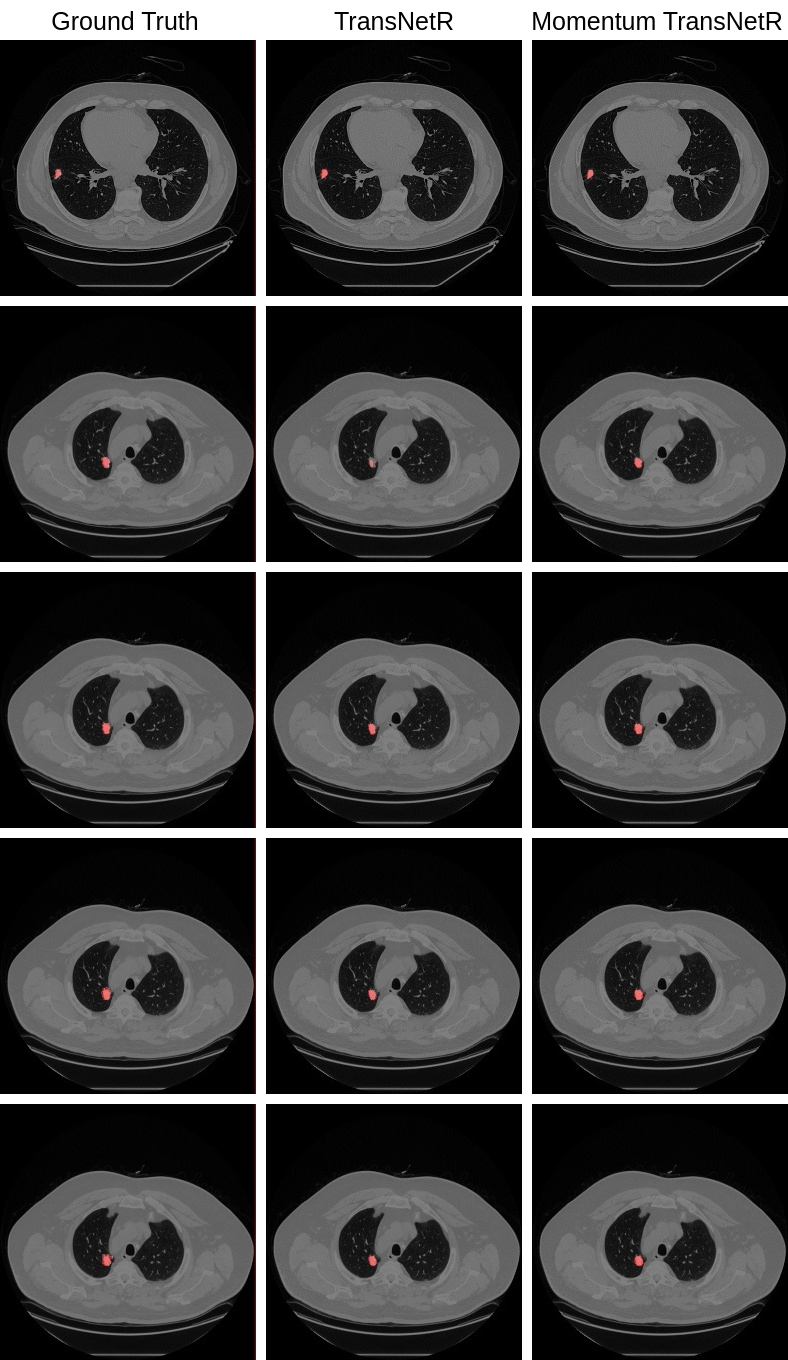}
    \caption{: Qualitative results of models trained Lung dataset on the TransNetR model. It can be observed that the Momentum-based method produces a more accurate segmentation map in all the cases.}
    \label{fig:enter-label1}
    \vspace{-7mm}
\end{figure}
\subsection{Medical Image Classification}
For the image classification task, we consider the RadImageNet~\cite{radimage} dataset, which is a medical imaging database that is publicly available and designed to improve transfer learning capabilities in medical imaging applications. It is one of the largest medical imaging classification datasets currently available and is intended for use by professionals in the field of healthcare. We conducted experiments on CT and MRI abdominal/pelvis using the entire dataset. The dataset comprises 28 disease classes, each with an average class size of 4994 and a total of 139,825 slices. The dataset is specifically designed to have slices per disease, although the overall scans are in 3D volumes. We conducted our experiments on the dataset of MRI images of the abdomen and pelvis. The dataset consisted of 26 different classes of diseases, with an average class size of 3513 slices and a total of 91,348 slices. Although there is some overlap between this dataset and the CT dataset, the MRI dataset has some unique disease classes, such as enlarged organs and liver disease, which are not found in the CT dataset. On the other hand, the CT dataset has a specific class for entire abdominal organs. 

\begin{table}[!htbp]
\scriptsize
\centering
\caption{Comparison of different segmentation models and our proposed momentum-based approach on liver cancer segmentation benchmark (LiTS) dataset.}
\label{tab:segmentationlITS3}
\begin{tabular}{c|c|c|c|c|c|c}
\toprule
\textbf{Model} & \textbf{mDSC} & \textbf{mIoU} & \textbf{Rec.} & \textbf{Prec.}  & \textbf{F2}  & \textbf{HD}  \\ \hline

ResUNet++~\cite{resunetplus}  & 73.82 & 70.63 & \textbf{76.10} & 91.13 &  72.21 & 1.23 \\
Momentum-ResUNet++ \textbf{(Ours)}  & \textbf{74.53}  & \textbf{71.88} & 74.98 & \textbf{95.46} & \textbf{72.83} & \textbf{1.22}   \\
\hline

ResUnet~\cite{resunet}  & 74.47 & 71.25 & 76.58 & 92.32 & 72.97 &  1.14\\
Momentum-ResUnet \textbf{(Ours)}  & \textbf{76.22}  & \textbf{72.79} & \textbf{77.45} & \textbf{92.76} & \textbf{74.10} &  \textbf{1.08}\\
\hline

TransNetR~\cite{transnetr}  & \textbf{78.74} &  75.18 & \textbf{78.16} & \textbf{95.59} & \textbf{76.86} &  \textbf{1.10}  \\
Momentum-TransNetR \textbf{(Ours)}  & 78.50  & \textbf{75.20} & 78.00 & 95.40 & 77.86 &  1.14\\
\hline

PVTFormer~\cite{pvt}  & 77.11 & 73.66 & 77.59 & \textbf{94.35} & 75.36 &  1.14  \\
Momentum-PVTFormer \textbf{(Ours)}  & \textbf{79.26}  & \textbf{75.67} & \textbf{80.74} & 93.87 & \textbf{77.87} &  \textbf{1.06}  \\
\bottomrule
\end{tabular}
\end{table}

\begin{table}[!htbp]
\scriptsize
\centering
\caption{Comparison of different segmentation models and our proposed momentum-based approach on healthy liver segmentation benchmark dataset.}
\label{tab:segmentationlITS2}
\begin{tabular}{c|c|c|c|c|c|c}
\toprule
\textbf{Model} & \textbf{mDSC} & \textbf{mIoU} & \textbf{Rec.} & \textbf{Prec.}  & \textbf{F2}  & \textbf{HD}  \\ \hline
ResUNet++~\cite{resunetplus}  & 85.70 & 77.80 & 79.78 & 97.01  & 81.96 & 3.61 \\
Momentum-ResUNet++ \textbf{(Ours)}  & \textbf{86.42}  & \textbf{78.90} & \textbf{80.57} & \textbf{97.14}  & \textbf{82.71} & \textbf{3.54} \\

\hline
ResUnet~\cite{resunet}  & 85.49 & 77.56 & 79.44 & 96.29  & 81.74 & 3.60 \\
Momentum-ResUnet \textbf{(Ours)} & \textbf{86.16}  & \textbf{78.45} & \textbf{80.59} & \textbf{96.49}  & \textbf{82.60} & \textbf{3.54} \\

\hline
TransNetR~\cite{transnetr}  & 85.77 & 78.44 & 79.42 & 96.49  & 81.78 & 3.58 \\
Momentum-TransNetR \textbf{(Ours)} & \textbf{86.99}  & \textbf{79.13} & \textbf{80.44} & \textbf{97.62}  & \textbf{82.84} & \textbf{3.55} \\

\hline

PVTFormer~\cite{pvt}  & 87.65 & 80.20 & \textbf{82.01} & 96.80 &  84.11 & \textbf{3.51} \\
Momentum-PVTFormer \textbf{(Ours)} &  \textbf{88.24} & \textbf{80.87} & 81.97 & \textbf{97.62}  &  \textbf{84.14} &  \textbf{3.51}\\
\bottomrule
\end{tabular}
\end{table}

\begin{table}[!htbp]
\scriptsize
\centering
\caption{Comparison of different segmentation models and our proposed momentum-based approach on Kvasir-Seg segmentation benchmark dataset.}
\label{tab:segmentationlITS}
\begin{tabular}{c|c|c|c|c|c|c}
\toprule
\textbf{Model} & \textbf{mDSC} & \textbf{mIoU} & \textbf{Rec.} & \textbf{Prec.}  & \textbf{F2}  & \textbf{HD}  \\ \hline

ResUNet++~\cite{resunetplus} & \textbf{66.56} & \textbf{56.35} & \textbf{73.25} & \textbf{71.70}  & \textbf{68.78}  &  \textbf{5.43}\\
Momentum-ResUNet++ \textbf{(Ours)} &  65.48 & 55.74 & 72.94 & 70.46 & 67.77 & 5.48\\
\hline

ResUNet~\cite{resunet}  & 78.32 & 68.85 &  81.33 & 83.42  &  79.40 &  4.79\\
Momentum-ResUNet \textbf{(Ours)} &  \textbf{79.89} & \textbf{70.20} & \textbf{82.45} & \textbf{83.29}  & \textbf{80.58} & \textbf{4.67}\\
\hline

TransNetR~\cite{transnetr}  & 87.89  & 81.08 & 88.53 & 91.22  & 87.89 & 4.07\\
Momentum-TransNetR \textbf{(Ours)}   & \textbf{88.50}  & \textbf{81.67} & \textbf{89.35} & \textbf{90.97}  & \textbf{88.43} & \textbf{4.06} \\
\hline

PVTFormer~\cite{pvt}  & 89.46  & 83.57 & 91.39 & 91.41  & 89.80 &  3.90\\
Momentum-PVTFormer \textbf{(Ours)} & \textbf{90.77} & \textbf{85.20} & \textbf{93.10} & \textbf{91.77}  & \textbf{91.47} &  \textbf{3.76}\\

\bottomrule
\end{tabular}
\end{table}


\begin{table*}[!htbp]
\scriptsize
    \centering
    \caption{Baseline models and the proposed method and their impact across 26 Classes in RedImageNet  Abdominal/Pelvis MRI Scans.}
    \begin{tabular}{c|c|c}
       \toprule
        {\textbf{Method}} & {\textbf{Accuracy}}  & {\textbf{MCC}} \\
        \hline
        MobileNet V2~\cite{mobilenetv2} & 82.94 & 47.12\\
        Momentum-MobileNet V2 \textbf{(Ours)} & \textbf{84.10} & \textbf{48.43}\\
        \hline
         ShuffleNet~\cite{shufflenet} &  84.56 & 55.40\\
        Momentum-ShuffleNet \textbf{(Ours)} & \textbf{85.10} & \textbf{55.99}\\
        \bottomrule
    \end{tabular}
    \label{MRIscantable1}
\end{table*}

We consider the Tensorflow-Keras~\cite{keras} framework to run the experiments, with MobileNet V2~\cite{mobilenetv2} and ShuffleNet~\cite{resnet} serving as baseline models classification networks. The networks are trained with a batch size of 32, an initial learning rate set at 0.00001, Adam~\cite{adam} optimizer, and a weight decay rate of $1e^{-4}$. The data is partitioned into three sets, with 80\% used for training, 10\% for validation, and 10\% for testing. The results obtained from MRI scan image data are presented in Table~\ref{MRIscantable1}, while those from the CT image data are presented in Table~\ref{CTscantable1}. All the classification experiments are conducted on an NVIDIA RTX 3090 GPU system.
\begin{table*}[!htbp]
\scriptsize
    \centering
    \caption{Baseline models and the proposed method and their impact on RadImageNet 28 classes Abdominal/Pelvis CT Scans.}
    \begin{tabular}{c|c|c}
       \toprule
        {\textbf{Method}} & {\textbf{Accuracy}}  & {\textbf{MCC}} \\
        \hline
        MobileNet V2~\cite{mobilenetv2} &  58.68 & 35.21\\
        Momentum-MobileNet V2 \textbf{(Ours)} & \textbf{60.79} & \textbf{37.89}\\
        \hline
         ShuffleNet~\cite{shufflenet} &  62.70 & 43.20\\
        Momentum-ShuffleNet \textbf{(Ours)} & \textbf{64.91} & \textbf{45.59}\\
        \bottomrule
    \end{tabular}
    \label{CTscantable1}
\end{table*}
\subsection{Performance Evaluation}
We have examined the momentum-based approach in various situations. We carried out experiments to explore the model's ability to learn on the test set of Liver, Lung, and Kvasir-SEG datasets for image segmentation and RadImageNet data for CT and MRI image classification. The outcomes pertaining to the observed dataset on the segmentation task have been presented in Table~\ref{tab:segmentationlITS1}, and Table~\ref{tab:segmentationlITS3}, and classification results are presented in Table~\ref{MRIscantable1} and Table~\ref{CTscantable1}. Our proposed methodology has yielded the most favourable results in terms of mIoU, dice score, precision, recall, F2, and HD score when compared to other models. For example, in the lung segmentation benchmark dataset, we got a 5.72\% improvement in dice score, a 5.04\% improvement in mIoU, an 8.02\% improvement in recall, a 4.42\% improvement in precision, a 6.43\% improvement in F2 score compared to the TransNetR model. In the medical image classification task, we consider the RadImageNet dataset. With our proposed method, in  Abdominal/pelvic CT scan data, we got a 2.11\% improvement on MobileNet V2 and a 2.21\% improvement on the ShuffleNet model.

Figure~\ref {fig:enter-label1} shows the outcomes of lung segmentation on the TransNetR model and the proposed method. It is noticeable that the proposed method has a higher segmentation accuracy in comparison to the state-of-the-art baselines. 

\section{Conclusion}

This study presents a novel momentum-based segmentation and classification approach that effectively segments liver, lung, and polyps using the momentum equation and residual block. The findings from various publicly available data demonstrate the efficacy of the proposed classification and segmentation approach. Based on a comprehensive comparison of our momentum algorithm on other datasets, our approach has consistently shown superior performance over our competitors. The quantitative and qualitative analysis results indicate that the momentum-based approach is more generalizable to most datasets, making it a suitable tool for clinical settings. Therefore, the proposed momentum-based approach provides a strong benchmark for developing algorithms that can assist clinicians in early lung, liver, and polyp detection and medical image classification.

\bibliographystyle{unsrtnat}
\bibliography{references}  






\end{document}